\documentclass[letterpaper]{article} 
\usepackage{aaai19m}  
\usepackage{times}  
\usepackage{helvet}  
\usepackage{courier}  
\usepackage{url}  
\usepackage{graphicx}  
\usepackage{subfigure}
\usepackage{booktabs}
\usepackage{bbding}
\usepackage{textcomp}
\usepackage{multirow}
\usepackage{amsmath}
\def\ie{{\em i.e.}}
\def\eg{{\em e.g.}}

\begin{document}
\title{Selective Refinement Network for High Performance Face Detection}
\author{
Cheng Chi$^{1,3*}$, Shifeng Zhang$^{2,3}$\thanks{These authors contributed equally to this work.}\thanks{Corresponding author}, Junliang Xing$^{2,3}$, Zhen Lei$^{2,3}$, Stan Z. Li$^{2,3}$, Xudong Zou$^{1,3}$\\
$^{1}$ Institute of Electronics, Chinese Academy of Sciences, Beijing, China\\
$^{2}$CBSR \& NLPR, Institute of Automation, Chinese Academy of Sciences, Beijing, China\\
$^{3}$University of Chinese Academy of Sciences, Beijing, China\\
{\tt\small chicheng15@mails.ucas.ac.cn, \{shifeng.zhang,jlxing,zlei,szli\}@nlpr.ia.ac.cn, xdzou@mail.ie.ac.cn}\\
}
\maketitle

\begin{abstract}
High performance face detection remains a very challenging problem, especially when there exists many tiny faces. This paper presents a novel single-shot face detector, named Selective Refinement Network (SRN), which introduces novel two-step classification and regression operations selectively into an anchor-based face detector to reduce false positives and improve location accuracy simultaneously. In particular, the SRN consists of two modules: the Selective Two-step Classification (STC) module and the Selective Two-step Regression (STR) module. The STC aims to filter out most simple negative anchors from low level detection layers to reduce the search space for the subsequent classifier, while the STR is designed to coarsely adjust the locations and sizes of anchors from high level detection layers to provide better initialization for the subsequent regressor. Moreover, we design a Receptive Field Enhancement (RFE) block to provide more diverse receptive field, which helps to better capture faces in some extreme poses. As a consequence, the proposed SRN detector achieves state-of-the-art performance on all the widely used face detection benchmarks, including AFW, PASCAL face, FDDB, and WIDER FACE datasets. Codes will be released to facilitate further studies on the face detection problem.
\end{abstract}

\section{Introduction}
Face detection is a long-standing problem in computer
vision with extensive applications including face alignment, face analysis, face recognition, etc. Starting from the pioneering work of Viola-Jones~\cite{DBLP:journals/ijcv/ViolaJ04}, face detection has made great progress. The performances on several well-known datasets have been improved consistently, even tend to be saturated. To further improve the performance of face detection has become a challenging issue. In our opinion, there remains room for improvement in two aspects: (a) \emph{recall efficiency}: number of false positives needs to be reduced at the high recall rates; (b) \emph{location accuracy}: accuracy of the bounding box location needs to be improved. These two problems are elaborated as follows.

\begin{figure}[t]
\centering
\subfigure[Effect on Class Imbalance]{
\label{fig:sc} 
\includegraphics[width=0.45\linewidth]{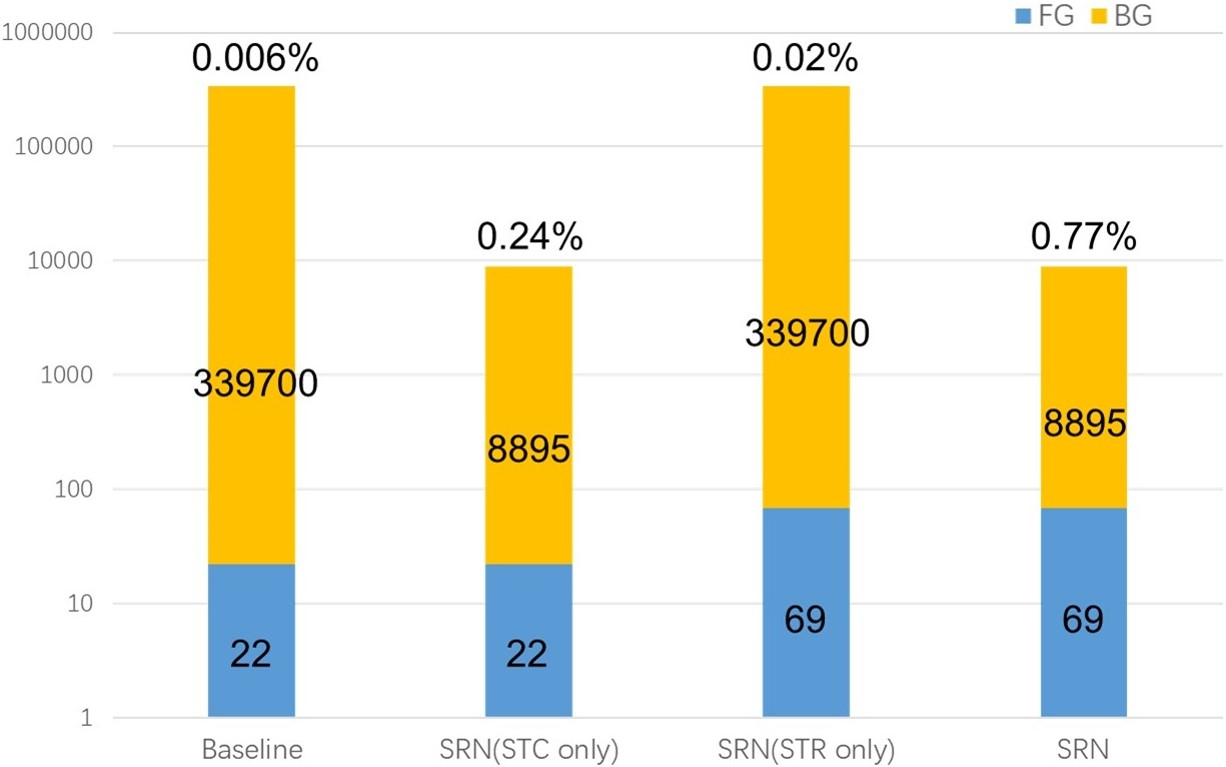}}
\subfigure[Recall Efficiency]{
\label{fig:prc} 
\includegraphics[width=0.45\linewidth]{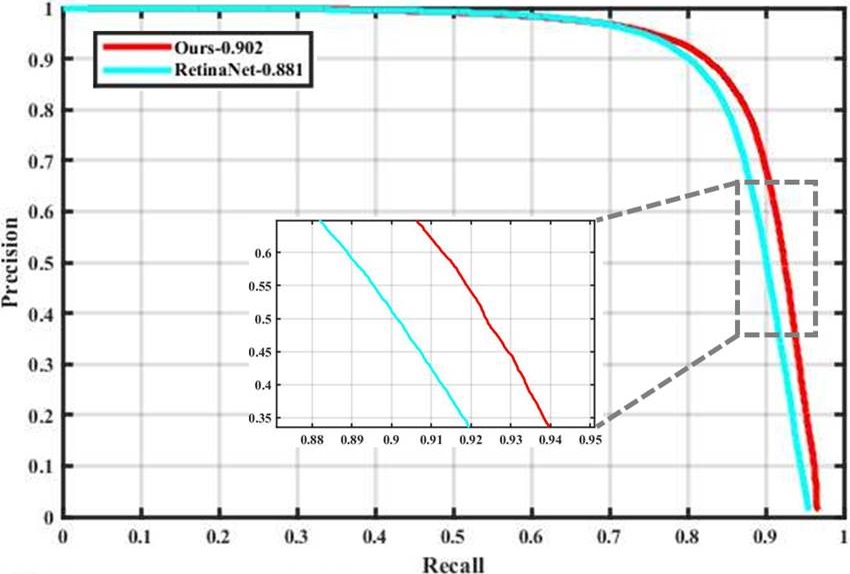}}
\subfigure[Adjusted Anchor]{
\label{fig:aa} 
\includegraphics[width=0.45\linewidth]{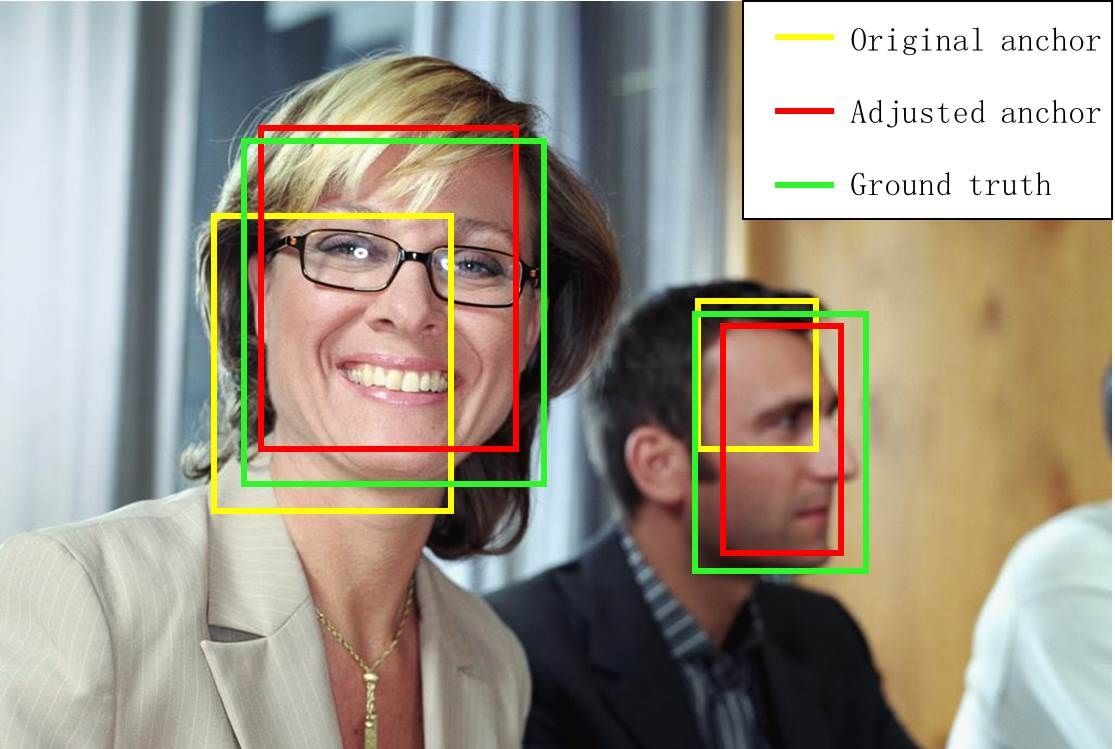}}
\subfigure[Location Accuracy]{
\label{fig:apc} 
\includegraphics[width=0.45\linewidth]{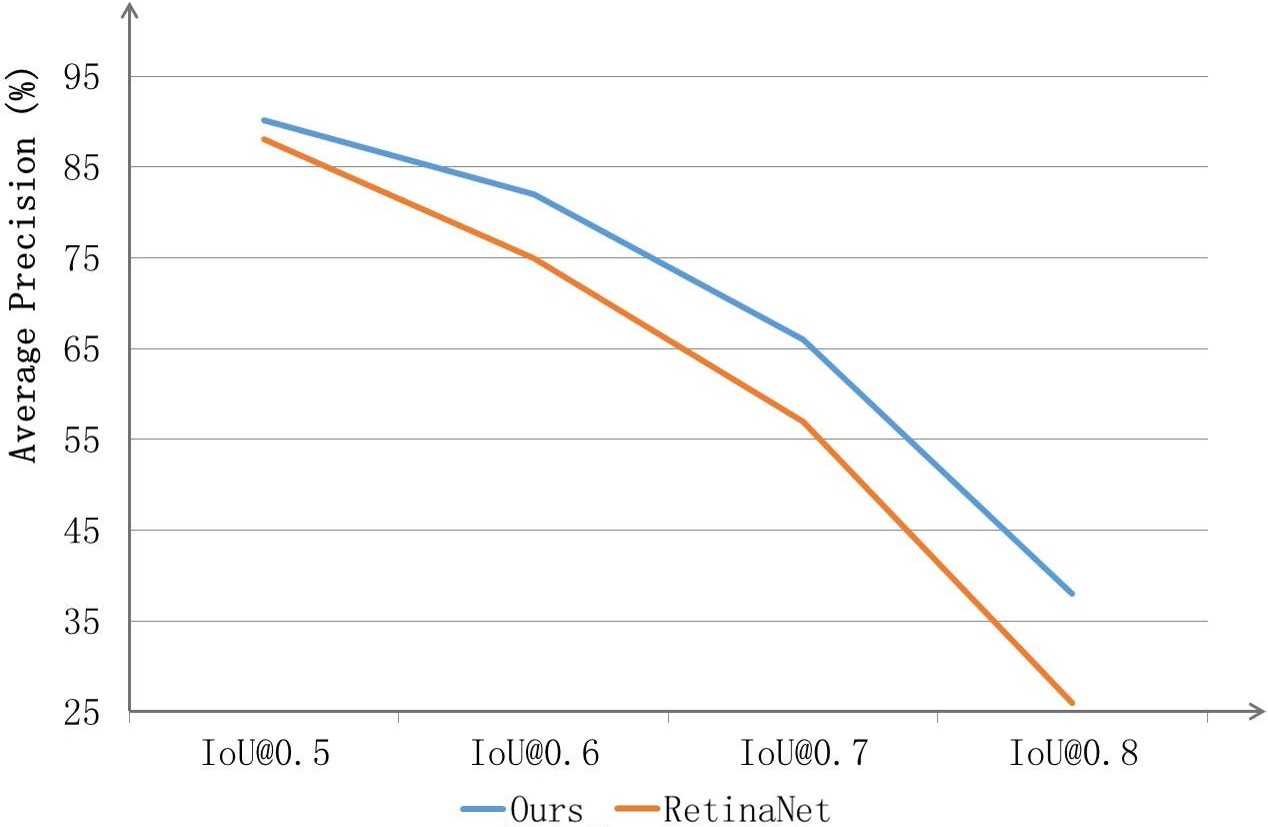}}
\vspace{-3.0mm}
\caption{The effects of STC and STR on recall efficiency and location accuracy. (a) The STC and STR increase the positives/negatives ratio by about $38$ and $3$ times respectively, (b) which improve the precision by about $20\%$ at high recall rates. (c) The STR provides better initialization for the subsequent regressor, (d) which produces more accurate locations, \ie, as the IoU threshold increases, the AP gap gradually increases.}
\vspace{-4.0mm}
\label{fig:effect-stc-str} 
\end{figure}

On the one hand, the average precision (AP) of current face detection algorithms is already very high, but the precision is not high enough at high recall rates, \eg, as shown in Figure \ref{fig:prc} of RetinaNet~\cite{DBLP:conf/iccv/LinPRK17}, the precision is only about $50\%$ (half of detections are false positives) when the recall rate is equal to $90\%$, which we define as the \emph{low recall efficiency}. Reflected on the shape of the Precision-Recall curve, it has extended far enough to the right, but not steep enough. The reason is that existing algorithms pay more attention to pursuing high recall rate but ignore the problem of excessive false positives. Analyzing with anchor-based face detectors, they detect faces by classifying and regressing a series of preset anchors, which are generated by regularly tiling a collection of boxes with different scales and aspect ratios. To detect the tiny faces, \eg, less than $16\times16$ pixels, it is necessary to tile plenty of small anchors over the image. This can improve the recall rate yet cause the the extreme class imbalance problem, which is the culprit leading to excessive false positives. To address this issue, researchers propose several solutions. R-CNN-like detectors~\cite{DBLP:conf/iccv/Girshick15,DBLP:journals/pami/RenHG017} address the class imbalance by a two-stage cascade and sampling heuristics. As for single-shot detectors, RetinaNet proposes the focal loss to focus training on a sparse set of hard examples and down-weight the loss assigned to well-classified examples. RefineDet~\cite{DBLP:journals/corr/abs-1711-06897} addresses this issue using a preset threshold to filter out negative anchors. However, RetinaNet takes all the samples into account, which also leads to quite a few false positives. Although RefineDet filters out a large number of simple negative samples, it uses hard negative mining in both two steps, and does not make full use of negative samples. Thus, the recall efficiency of them both can be improved.

On the other hand, the location accuracy in the face detection task is gradually attracting the attention of researchers. Although current evaluation criteria of most face detection datasets~\cite{fddbTech,DBLP:conf/cvpr/YangLLT16} do not focus on the location accuracy, the WIDER Face Challenge\footnote{http://wider-challenge.org} adopts MS COCO~\cite{DBLP:conf/eccv/LinMBHPRDZ14} evaluation criterion, which puts more emphasis on bounding box location accuracy. To visualize this issue, we use different IoU thresholds to evaluate our trained face detector based on RetinaNet on the WIDER FACE dataset. As shown in Figure \ref{fig:apc}, as the IoU threshold increases, the AP drops dramatically, indicating that the accuracy of the bounding box location needs to be improved. To this end, Gidaris et al.~\cite{DBLP:conf/iccv/GidarisK15} propose iterative regression during inference to improve the accuracy. Cascade R-CNN~\cite{DBLP:journals/corr/abs-1712-00726} addresses this issue by cascading R-CNN with different IoU thresholds. RefineDet~\cite{DBLP:journals/corr/abs-1711-06897} applies two-step regression to single-shot detector. However, blindly adding multi-step regression to the specific task (\ie, face detection) is often counterproductive.

\begin{figure*}[ht!]
\centering
\includegraphics[width=0.95\linewidth]{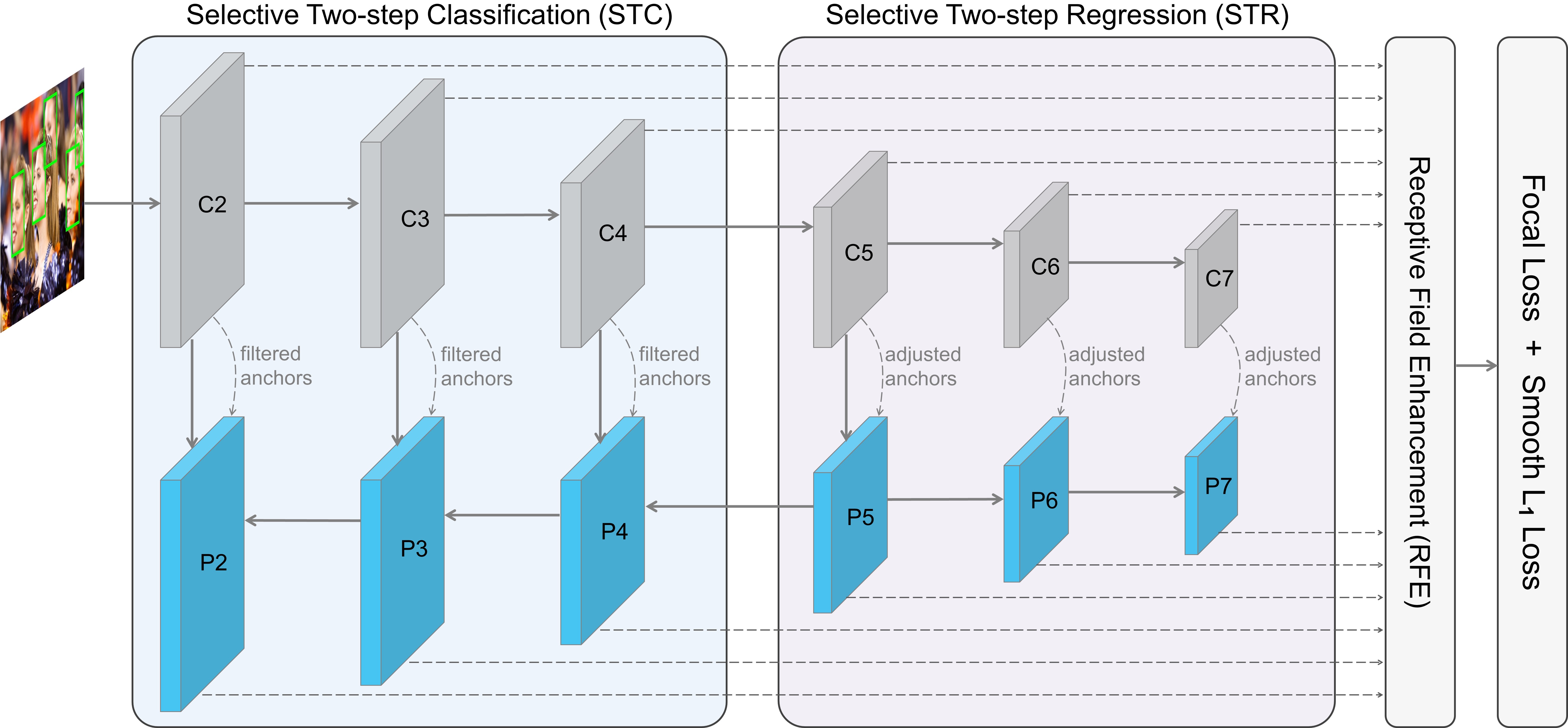}
\caption{Network structure of SRN. It consists of STC, STR, and RFB. STC uses the first-step classifier to filter out most simple negative anchors from low level detection layers to reduce the search space for the second-step classifier. STR applies the first-step regressor to coarsely adjust the locations and sizes of anchors from high level detection layers to provide better initialization for the second-step regressor. RFE provides more diverse receptive fields to better capture extreme-pose faces.
}
\label{fig:framework}
\end{figure*}

In this paper, we investigate the effects of two-step classification and regression on different levels of detection layers and propose a novel face detection framework, named Selective Refinement Network (SRN), which selectively applies two-step classification and regression to specific levels of detection layers. The network structure of SRN is shown in Figure \ref{fig:framework}, which consists of two key modules, named as the Selective Two-step Classification (STC) module and the Selective Two-step Regression (STR) module. Specifically, the STC is applied to filter out most simple negative samples (illustrated in Figure \ref{fig:sc}) from the low levels of detection layers, which contains $88.9\%$ samples. As shown in Figure \ref{fig:prc}, RetinaNet with STC improves the recall efficiency to a certain extent. On the other hand, the design of STR draws on the cascade idea to coarsely adjust the locations and sizes of anchors (illustrated in Figure \ref{fig:aa}) from high levels of detection layers to provide better initialization for the subsequent regressor. In addition, we design a Receptive Field Enhancement (RFE) to provide more diverse receptive fields to better capture the extreme-pose faces. Extensive experiments have been conducted on AFW, PASCAL face, FDDB, and WIDER FACE benchmarks and we set a new state-of-the-art performance.

In summarization, we have made the following main contributions to the face detection studies:
\begin{itemize}
\item We present a STC module to filter out most simple negative samples from low level layers to reduce the classification search space.
\item We design a STR module to coarsely adjust the locations and sizes of anchors from high level layers to provide better initialization for the subsequent regressor.
\item We introduce a RFE module to provide more diverse receptive fields for detecting extreme-pose faces.
\item We achieve state-of-the-art results on AFW, PASCAL face, FDDB, and WIDER FACE datasets.
\end{itemize}

\section{Related Work}
Face detection has been a challenging research field since its emergence in the 1990s. Viola and Jones pioneer to use Haar features and AdaBoost to train a face detector with promising accuracy and efficiency~\cite{DBLP:journals/ijcv/ViolaJ04}, which inspires several different approaches afterwards~\cite{DBLP:journals/pami/LiaoJL16,DBLP:journals/ijcv/BrubakerWSMR08,DBLP:conf/iccv/PhamC07}. Apart from those, another important job is the introduction of Deformable Part Model (DPM)~\cite{DBLP:conf/eccv/MathiasBPG14,DBLP:conf/cvpr/YanLWL14,DBLP:conf/cvpr/ZhuR12}. 

Recently, face detection has been dominated by the CNN-based methods. 
CascadeCNN~\cite{DBLP:conf/cvpr/LiLSBH15} improves detection accuracy by training a serious of interleaved CNN models and following work~\cite{DBLP:conf/cvpr/QinYLH16} proposes to jointly train the cascaded CNNs to realize end-to-end optimization. MTCNN~\cite{DBLP:journals/spl/ZhangZLQ16} proposes a joint face detection and alignment method using multi-task cascaded CNNs. Faceness~\cite{DBLP:conf/iccv/YangLLT15} formulates face detection as scoring facial parts responses to detect faces under severe occlusion. UnitBox~\cite{DBLP:conf/mm/YuJWCH16} introduces an IoU loss for bounding box prediction. EMO~\cite{zhu2018seeing} proposes an Expected Max Overlapping score to evaluate the quality of anchor matching. SAFD~\cite{hao2017scale} develops a scale proposal stage which automatically normalizes face sizes prior to detection. S$^{2}$AP~\cite{song2018beyond} pays attention to specific scales in image pyramid and valid locations in each scales layer. PCN~\cite{shi2018real} proposes a cascade-style structure to rotate faces in a coarse-to-fine manner. Recent work~\cite{bai2018finding} designs a novel network to directly generate a clear super-resolution face from a blurry small one.

Additionally, face detection has inherited some achievements from generic object detectors, such as Faster R-CNN~\cite{DBLP:journals/pami/RenHG017}, SSD~\cite{DBLP:conf/eccv/LiuAESRFB16}, FPN~\cite{DBLP:conf/cvpr/LinDGHHB17} and RetinaNet~\cite{DBLP:conf/iccv/LinPRK17}. 
Face R-CNN~\cite{wang2017face} combines Faster R-CNN with hard negative mining and achieves promising results. 
Face R-FCN~\cite{wang2017detecting} applies R-FCN in face detection and makes according improvements.
The face detection model for finding tiny faces~\cite{DBLP:conf/cvpr/HuR17} trains separate detectors for different scales. S$^{3}$FD~\cite{DBLP:conf/iccv/abs-1708-05237} presents multiple strategies onto SSD to compensate for the matching problem of small faces. SSH~\cite{DBLP:conf/iccv/NajibiSCD17} models the context information by large filters on each prediction module. PyramidBox~\cite{tang2018pyramidbox} utilizes contextual information with improved SSD network structure.
FAN~\cite{wang2017fan} proposes an anchor-level attention into RetinaNet to detect the occluded faces. In this paper, inspired by the multi-step classification and regression in RefineDet~\cite{DBLP:journals/corr/abs-1711-06897} and the focal loss in RetinaNet, we develop a state-of-the-art face detector.

\section{Selective Refinement Network}

\subsection{Network Structure}
The overall framework of SRN is shown in Figure \ref{fig:framework}, we describe each component as follows.

{\flushleft \textbf{Backbone.} }
We adopt ResNet-50~\cite{DBLP:conf/cvpr/HeZRS16} with 6-level feature pyramid structure as the backbone network for SRN. The feature maps extracted from those four residual blocks are denoted as C2, C3, C4, and C5, respectively. C6 and C7 are just extracted by two simple down-sample $3\times3$ convolution layers after C5. The lateral structure between the bottom-up and the top-down pathways is the same as~\cite{DBLP:conf/cvpr/LinDGHHB17}. P2, P3, P4, and P5 are the feature maps extracted from lateral connections, corresponding to C2, C3, C4, and C5 that are respectively of the same spatial sizes, while P6 and P7 are just down-sampled by two $3\times3$ convolution layers after P5.

{\flushleft \textbf{Dedicated Modules.} }
The STC module selects C2, C3, C4, P2, P3, and P4 to perform two-step classification, while the STR module selects C5, C6, C7, P5, P6, and P7 to conduct two-step regression. The RFE module is responsible for enriching the receptive field of features that are used to predict the classification and location of objects.

{\flushleft \textbf{Anchor Design.} }
At each pyramid level, we use two specific scales of anchors (\ie, $2S$ and $2\sqrt{2}S$, where $S$ represents the total stride size of each pyramid level) and one aspect ratios (\ie, $1.25$). In total, there are $A=2$ anchors per level and they cover the scale range $8-362$ pixels across levels with respect to the network's input image.

{\flushleft \textbf{Loss Function.} }
We append a hybrid loss at the end of the deep architecture, which leverage the merits of the focal loss and the smooth L$_1$ loss to drive the model to focus on more hard training examples and learn better regression results. 

\subsection{Selective Two-Step Classification}
Introduced in RefineDet~\cite{DBLP:journals/corr/abs-1711-06897}, the two-step classification is a kind of cascade classification implemented through a two-step network architecture, in which the first step filters out most simple negative anchors using a preset negative threshold $\theta=0.99$ to reduce the search space for the subsequent step. For anchor-based face detectors, it is necessary to tile plenty of small anchors over the image to detect small faces, which causes the extreme class imbalance between the positive and negative samples. For example, in the SRN structure with the $1024\times1024$ input resolution, if we tile $2$ anchors at each anchor point, the total number of samples will reach $300k$. Among them, the number of positive samples is only a few dozen or less. To reduce search space of classifier, it is essential to do two-step classification to reduce the false positives.

However, it is unnecessary to perform two-step classification in all pyramid levels. Since the anchors tiled on the three higher levels (\ie, P5, P6, and P7) only account for $11.1\%$ and the associated features are much more adequate. Therefore, the classification task is relatively easy in these three higher pyramid levels. It is thus dispensable to apply two-step classification on the three higher pyramid levels, and if applied, it will lead to an increase in computation cost. In contrast, the three lower pyramid levels (\ie, P2, P3, and P4) have the vast majority of samples ($88.9\%$) and lack of adequate features. It is urgently needed for these low pyramid levels to do two-step classification in order to alleviate the class imbalance problem and reduce the search space for the subsequent classifier.

Therefore, our STC module selects C2, C3, C4, P2, P3, and P4 to perform two-step classification. As the statistical result shown in Figure \ref{fig:sc}, the STC increases the positive/negative sample ratio by approximately $38$ times, from around $1$:$15441$ to $1$:$404$. In addition, we use the focal loss in both two steps to make full use of samples. Unlike RefineDet~\cite{DBLP:journals/corr/abs-1711-06897}, the SRN shares the same classification module in the two steps, since they have the same task to distinguish the face from the background. The experimental results of applying the two-step classification on each pyramid level are shown in Table \ref{tab:stc_per_level}. Consistent with our analysis, the two-step classification on the three lower pyramid levels helps to improve performance, while on the three higher pyramid levels is ineffective.

The loss function for STC consists of two parts, \ie, the loss in the first step and the second step. For the first step, we calculate the focal loss for those samples selected to perform two-step classification. And for the second step, we just focus on those samples that remain after the first step filtering. With these definitions, we define the loss function as:
\begin{equation}
\begin{aligned}
{\cal L}_\text{STC} (\{p_i\},\{q_i\})=\frac{1}{N_{\text{s}_1}}  \sum_{i\in \Omega}{\cal L}_{\text{FL}}(p_i,l_i^\ast) \\
+ \frac{1}{N_{\text{s}_2}}  \sum_{i\in \Phi}{\cal L}_{\text{FL}}(q_i, l_i^\ast),
\end{aligned}
\label{1}
\end{equation}
where $i$ is the index of anchor in a mini-batch, $p_i$ and $q_i$ are the predicted confidence of the anchor $i$ 
being a face in the first and second steps, $l_i^\ast$ is the ground truth class label of anchor $i$, $N_{\text{s1}}$ and $N_{\text{s2}}$ are the numbers of positive anchors in the first and second steps, $\Omega$ represents a collection of samples selected for two-step classification, and $\Phi$ represents a sample set that remains after the first step filtering. The binary classification loss ${\cal L}_{\text{FL}}$ is the sigmoid focal loss over two classes (face {\em vs.} background).

\subsection{Selective Two-Step Regression}
In the detection task, to make the location of bounding boxes more accurate has always been a challenging problem. Current one-stage methods rely on one-step regression based on various feature layers, which is inaccurate in some challenging scenarios, \eg, MS COCO-style evaluation standard. In recent years, using cascade structure \cite{DBLP:journals/corr/abs-1711-06897,DBLP:journals/corr/abs-1712-00726} to conduct multi-step regression is an effective method to improve the accuracy of the detection bounding boxes.

However, blindly adding multi-step regression to the specific task (\ie, face detection) is often counterproductive. Experimental results (see Table \ref{tab:str_per_level}) indicate that applying two-step regression in the three lower pyramid levels impairs the performance. The reasons behind this phenomenon are twofold: 1) the three lower pyramid levels are associated with plenty of small anchors to detect small faces. These small faces are characterized by very coarse feature representations, so it is difficult for these small anchors to perform two-step regression; 2) in the training phase, if we let the network pay too much attention to the difficult regression task on the low pyramid levels, it will cause larger regression loss and hinder the more important classification task. 

Based on the above analyses, we selectively perform two-step regression on the three higher pyramid levels. The motivation behind this design is to sufficiently utilize the detailed features of large faces on the three higher pyramid levels to regress more accurate locations of bounding boxes and to make three lower pyramid levels pay more attention to the classification task. This divide-and-conquer strategy makes the whole framework more efficient.

The loss function of STR also consists of two parts, which is shown as below:
\begin{equation}
\begin{aligned}
{\cal L}_\text{STR}(\{x_i\},\{t_i\})=\sum_{i\in \Psi}[l_i^\ast=1]{\cal L}_{\text{r}}(x_i, g_i^\ast) \\
+ \sum_{i\in \Phi}[l_i^\ast=1]{\cal L}_{\text{r}}(t_i, g_i^\ast),
\end{aligned}
\label{1}
\end{equation}
where $g_i^\ast$ is the ground truth location and size of anchor $i$, $x_i$ is the refined coordinates of the anchor $i$ in the first step, $t_i$ is the coordinates of the bounding box in the second step, $\Psi$ represents a collection of samples selected for two-step regression, $l_i^\ast$ and $\Phi$ are the same as defined in STC. Similar to Faster R-CNN \cite{DBLP:journals/pami/RenHG017}, we use the smooth L$_1$ loss as the regression loss $L_{\text{r}}$. The Iverson bracket indicator function $[l_i^\ast=1]$ outputs $1$ when the condition is true, \ie, $l_i^\ast=1$ (the anchor is not the negative), and $0$ otherwise. Hence $[l_i^\ast=1]{\cal L}_{\text{r}}$ indicates that the regression loss is ignored for negative anchors.

\subsection{Receptive Field Enhancement}
At present, most detection networks utilize ResNet and VGGNet as the basic feature extraction module, while both of them possess square receptive fields. The singleness of the receptive field affects the detection of objects with different aspect ratios. This issue seems unimportant in face detection task, because the aspect ratio of face annotations is about $1$:$1$ in many datasets. Nevertheless, statistics shows that the WIDER FACE training set has a considerable part of faces that have an aspect ratio of more than $2$ or less than $0.5$. Consequently, there is mismatch between the receptive field of network and the aspect ratio of faces.

\begin{figure}[t!]
\centering
\includegraphics[width=0.7\linewidth]{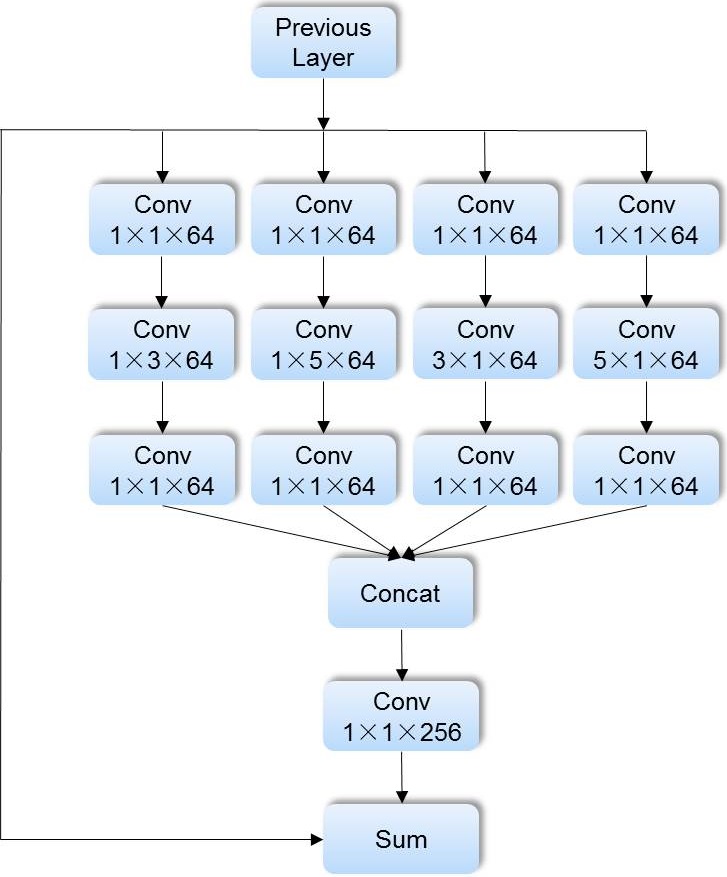}
\caption{Structure of RFE module.}
\label{fig:rfe}
\end{figure}

To address this issue, we propose a module named Receptive Field Enhancement (RFE) to diversify the receptive field of features before predicting classes and locations. In particular, RFE module replaces the middle two convolution layers in the class subnet and the box subnet of RetinaNet. The structure of RFE is shown in Figure \ref{fig:rfe}. Our RFE module adopts a four-branch structure, which is inspired by the Inception block \cite{DBLP:conf/cvpr/SzegedyLJSRAEVR15}. To be specific, first, we use a $1\times1$ convolution layer to decrease the channel number to one quarter of the previous layer. Second, we use $1\times k$ and $k\times 1$ ($k=3$ and $5$) convolution layer to provide rectangular receptive field. Through another $1\times1$ convolution layer, the feature maps from four branches are concatenated together. Additionally, we apply a shortcut path to retain the original receptive field from previous layer.

\section{Training and Inference}

{\flushleft \textbf{Training Dataset.} }
All the models are trained on the training set of the WIDER FACE dataset~\cite{DBLP:conf/cvpr/YangLLT16}. It consists of $393,703$ annotated face bounding boxes in $32,203$ images with variations in pose, scale, facial expression, occlusion, and lighting condition. The dataset is split into the training ($40\%$), validation ($10\%$) and testing ($50\%$) sets, and defines three levels of difficulty: Easy, Medium, Hard, based on the detection rate of EdgeBox~\cite{DBLP:conf/eccv/ZitnickD14}.

{\flushleft \textbf{Data Augmentation.} }
To prevent over-fitting and construct a robust model, several data augmentation strategies are used to adapt to face variations, described as follows.
\begin{itemize}
\item[1)] Applying some photometric distortions introduced in previous work~\cite{DBLP:journals/corr/Howard13} to the training images.
\item[2)] Expanding the images with a random factor in the interval $[1,2]$ by the zero-padding operation.
\item[3)] Cropping two square patches and randomly selecting one for training. One patch is with the size of the image's shorter side and the other one is with the size determined by multiplying a random number in the interval $[0.5,1.0]$ by the image's shorter side.
\item[4)] Flipping the selected patch randomly and resizing it to $1024\times1024$ to get the final training sample.
\end{itemize}

{\noindent \textbf{Anchor Matching.} }
During the training phase, anchors need to be divided into positive and negative samples. Specifically, anchors are assigned to ground-truth face boxes using an intersection-over-union (IoU) threshold of $\theta_{p}$; and to background if their IoU is in $[0, \theta_{n})$. If an anchor is unassigned, which may happen with overlap in $[\theta_{n}, \theta_{p})$, it is ignored during training. Empirically, we set $\theta_{n}=0.3$ and $\theta_{p}=0.7$ for the first step, and $\theta_{n}=0.4$ and $\theta_{p}=0.5$ for the second step. 

{\flushleft \textbf{Optimization.} }
The loss function for SRN is just the sum of the STC loss and the STR loss, \ie, ${\cal L} = {\cal L}_\text{STC} + {\cal L}_\text{STR}$. The backbone network is initialized by the pretrained ResNet-50 model~\cite{DBLP:journals/ijcv/RussakovskyDSKS15} and all the parameters in the newly added convolution layers are initialized by the ``xavier" method. We fine-tune the SRN model using SGD with $0.9$ momentum, $0.0001$ weight decay, and batch size $32$. We set the learning rate to $10^{-2}$ for the first $100$ epochs, and decay it to $10^{-3}$ and $10^{-4}$ for another $20$ and $10$ epochs, respectively. We implement SRN using the PyTorch library~\cite{paszke2017pytorch}.

{\flushleft \textbf{Inference.} }
In the inference phase, the STC first filters  the regularly tiled anchors on the selected pyramid levels with the negative confidence scores larger than the threshold $\theta=0.99$, and then STR adjusts the locations and sizes of selected anchors. After that, the second step takes over these refined anchors, and outputs top $2000$ high confident detections. Finally, we apply the non-maximum suppression (NMS) with jaccard overlap of $0.5$ to generate the top $750$ high confident detections per image as the final results.

\begin{figure*}[t]
\centering
\subfigure[AFW]{
\label{fig:AFW}
\includegraphics[width=0.325\linewidth]{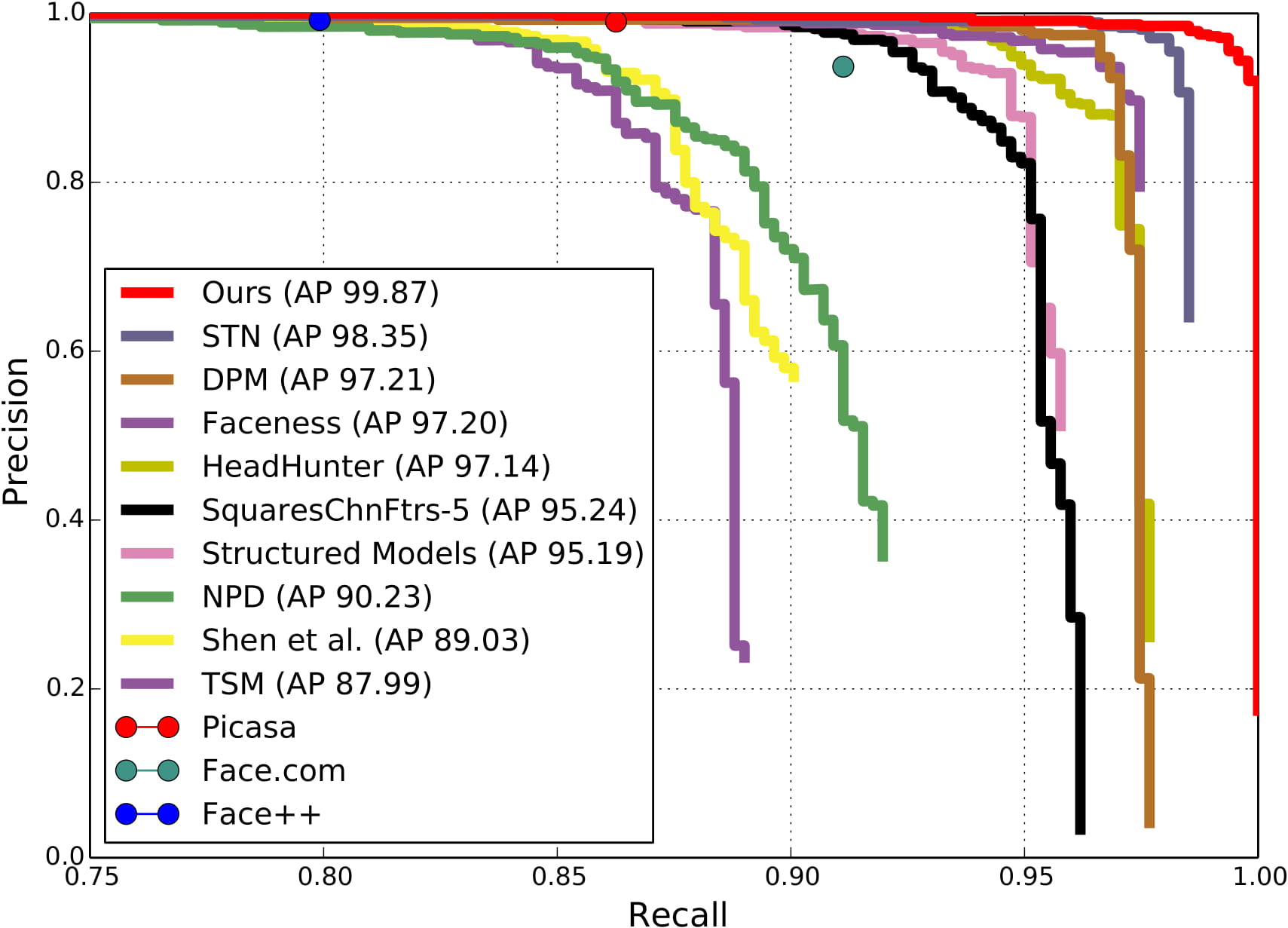}}
\subfigure[PASCAL face]{
\label{fig:PASCAL}
\includegraphics[width=0.325\linewidth]{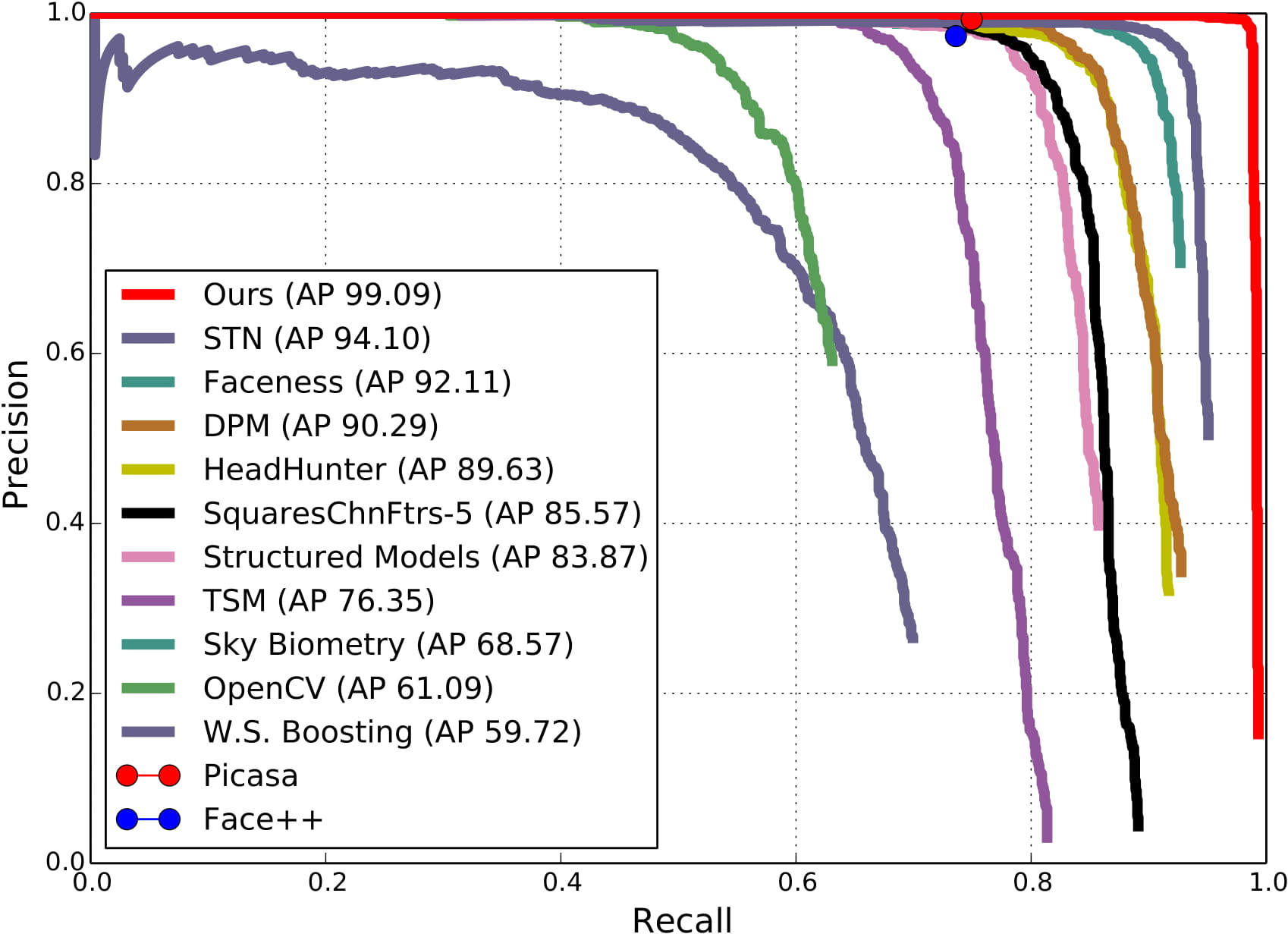}}
\subfigure[FDDB]{
\label{fig:FDDB}
\includegraphics[width=0.325\linewidth]{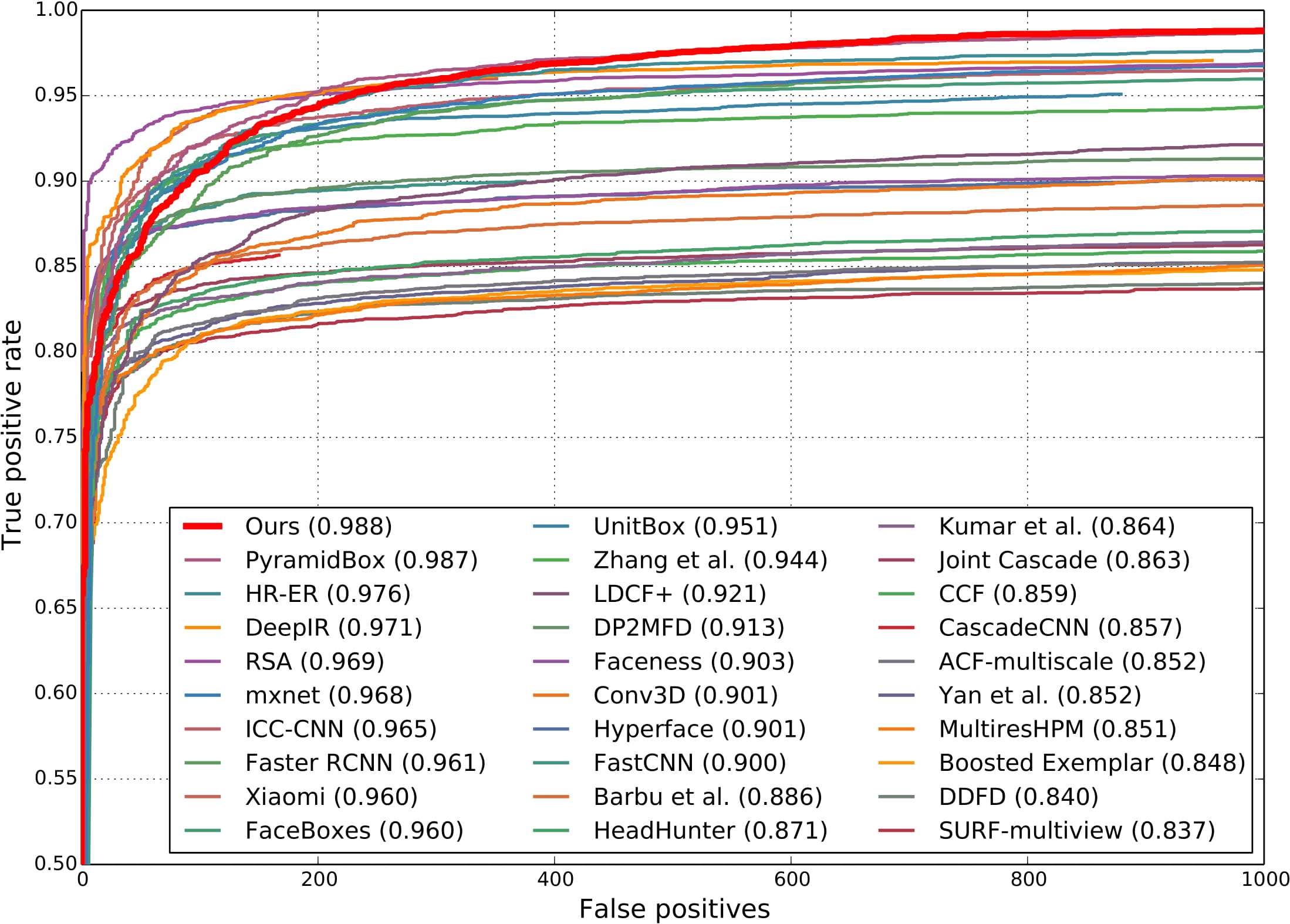}}
\caption{Evaluation on the common face detection datasets.}
\label{fig:evaluation}
\end{figure*}

\section{Experiments}

We first analyze the proposed method in detail to verify the effectiveness of our contributions. Then we evaluate the final model on the common face detection benchmark datasets, including AFW \cite{DBLP:conf/cvpr/ZhuR12}, PASCAL Face \cite{DBLP:journals/ivc/YanZLL14}, FDDB \cite{fddbTech}, and WIDER FACE \cite{DBLP:conf/cvpr/YangLLT16}.

\subsection{Model Analysis}

We conduct a set of ablation experiments on the WIDER FACE dataset to analyze our model in detail. For a fair comparison, we use the same parameter settings for all the experiments, except for specified changes to the components. All models are trained on the WIDER FACE training set and evaluated on the validation set.

{\flushleft \textbf{Ablation Setting.} }
To better understand SRN, we ablate each component one after another to examine how each proposed component affects the final performance. Firstly, we use the ordinary prediction head in \cite{DBLP:conf/iccv/LinPRK17} instead of the proposed RFE. Secondly, we ablate the STR or STC module to verity their effectiveness. The results of ablation experiments are listed in Table \ref{tab:ablation} and some promising conclusions can be drawn as follows.

\begin{table}[h]
\centering
\setlength{\tabcolsep}{9.0pt}
\caption{Effectiveness of various designs on the AP performance.}
\footnotesize{
\begin{tabular}{c|ccccc}
\toprule[0.5pt]
\multicolumn{1}{c|}{Component} & \multicolumn{4}{c}{SRN}\\
\hline
STC & & \Checkmark & & \Checkmark & \Checkmark \\
STR & & & \Checkmark & \Checkmark & \Checkmark \\
RFE & & & & & \Checkmark \\
\hline
{\em Easy} subset & 95.1 & 95.3 & 95.9 & 96.1 &\textbf{96.4}\\
{\em Medium} subset & 93.9 & 94.4 & 94.8 & 95.0 &\textbf{95.3}\\
{\em Hard} subset & 88.0 & 89.4 & 88.8 & 90.1 &\textbf{90.2}\\
\bottomrule[1.5pt]
\end{tabular}}
\label{tab:ablation}
\end{table}

{\flushleft \textbf{Selective Two-step Classification.} }
Experimental results of applying two-step classification to each pyramid level are shown in Table \ref{tab:stc_per_level}, indicating that applying two-step classification to the low pyramid levels improves the performance, especially on tiny faces. Therefore, the STC module selectively applies the two-step classification on the low pyramid levels (\ie, P2, P3, and P4), since these levels are associated with lots of small anchors, which are the main source of false positives. As shown in Table \ref{tab:ablation}, we find that after using the STC module, the AP scores of the detector are improved from $95.1\%$, $93.9\%$ and $88.0\%$ to $95.3\%$, $94.4\%$ and $89.4\%$ on the Easy, Medium and Hard subsets, respectively. In order to verify whether the improvements benefit from reducing the false positives, we count the number of false positives under different recall rates. As listed in Table \ref{tab:fp_num}, our STC effectively reduces the false positives across different recall rates, demonstrating the effectiveness of the STC module.

\vspace{-1.5mm}
\begin{table}[h]
\centering
\setlength{\tabcolsep}{3pt}
\caption{AP performance of the two-step classification applied to each pyramid level.}
\setlength{\tabcolsep}{5.2pt}
\begin{tabular}{c|c|cccccc}
\toprule[1.5pt]
STC & B & P2 & P3 & P4 & P5 & P6 & P7 \\
\hline
{\em Easy} & 95.1 & \bf 95.2 & \bf 95.2 & \bf 95.2 & 95.0 & 95.1 & 95.0 \\
{\em Medium} & 93.9 & \bf 94.2 & \bf 94.3 & \bf 94.1 & 93.9 & 93.7 & 93.9 \\
{\em Hard} & 88.0 & \bf 88.9 & \bf 88.7 & \bf 88.5 & 87.8 & 88.0 & 87.7 \\
\bottomrule[1.5pt]
\end{tabular}
\vspace{-5mm}
\label{tab:stc_per_level}
\end{table}

\begin{table}[h]
\centering
\setlength{\tabcolsep}{3pt}
\caption{Number of false positives at different recall rates.}
\setlength{\tabcolsep}{2.0pt}
\begin{tabular}{c|cccccc}
\toprule[1.5pt]
Recall ($\%$) & 10 & 30 & 50 & 80 & 90 & 95 \\
\hline
$\#$ FP of RetinaNet & 3 & 24 & 126 & 2801 & 27644 & 466534\\
$\#$ FP of SRN (STC only) & 1 & 20 & 101 & 2124 & 13163 & 103586\\
\bottomrule[1.5pt]
\end{tabular}
\vspace{-1.5mm}
\label{tab:fp_num}
\end{table}

\begin{figure*}[h]
\centering
\subfigure[Val: Easy]{
\label{fig:ve} 
\includegraphics[width=0.325\linewidth]{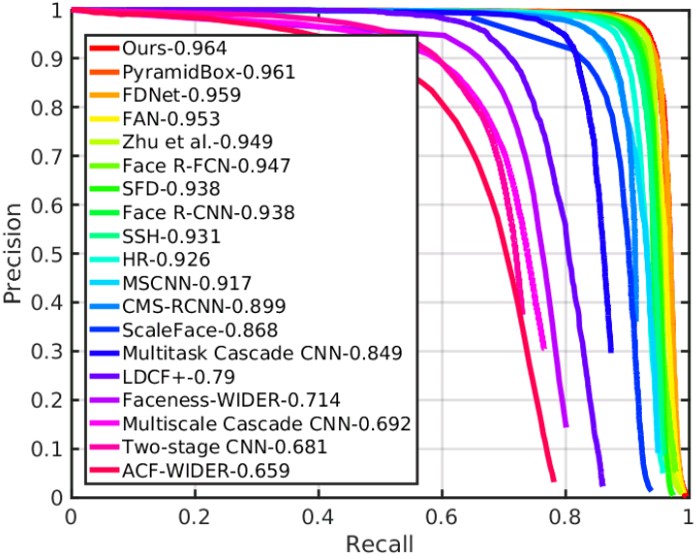}}
\subfigure[Val: Medium]{
\label{fig:vm} 
\includegraphics[width=0.325\linewidth]{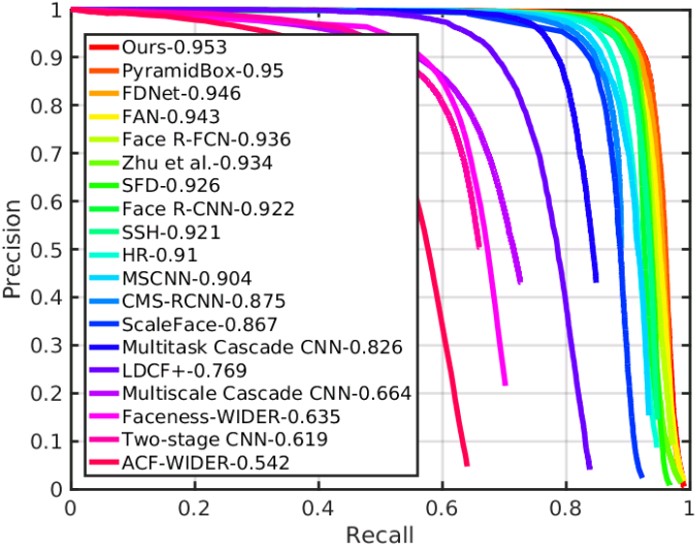}}
\subfigure[Val: Hard]{
\label{fig:vh} 
\includegraphics[width=0.325\linewidth]{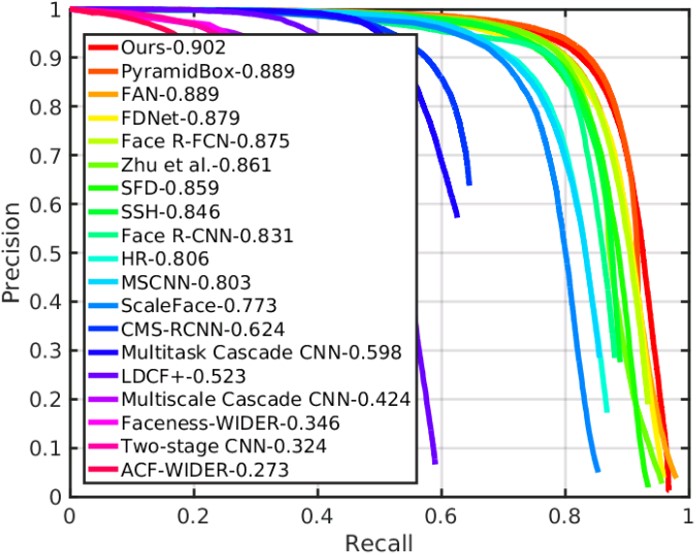}}
\subfigure[Test: Easy]{
\label{fig:te} 
\includegraphics[width=0.325\linewidth]{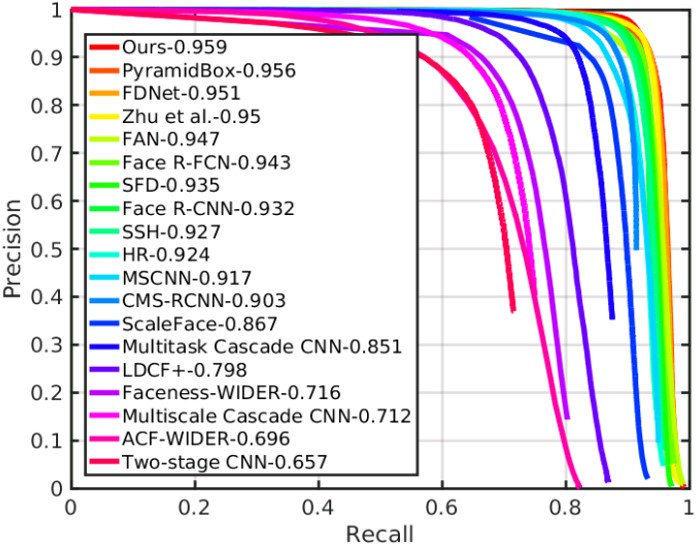}}
\subfigure[Test: Medium]{
\label{fig:tm} 
\includegraphics[width=0.325\linewidth]{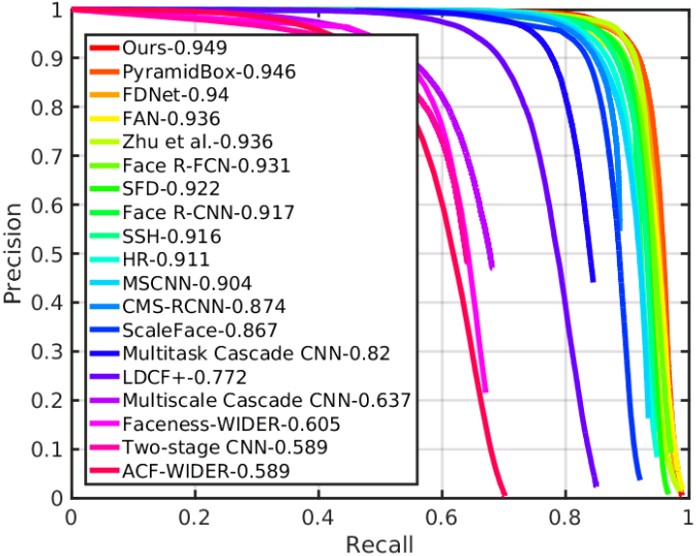}}
\subfigure[Test: Hard]{
\label{fig:th} 
\includegraphics[width=0.325\linewidth]{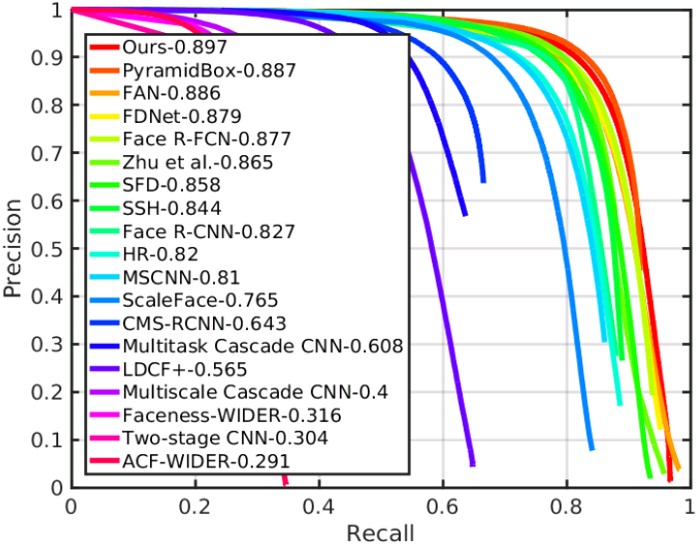}}
\caption{Precision-recall curves on WIDER FACE validation and testing subsets.}
\label{fig:wider-face-ap} 
\end{figure*}

{\flushleft \textbf{Selective Two-step Regression.} }
We only add the STR module to our baseline detector to verify its effectiveness. As shown in Table \ref{tab:ablation}, it produces much better results than the baseline, with $0.8\%$, $0.9\%$ and $0.8\%$ AP improvements on the Easy, Medium, and Hard subsets. Experimental results of applying two-step regression to each pyramid level (see Table \ref{tab:str_per_level}) confirm our previous analysis. Inspired by the detection evaluation metric of MS COCO, we use $4$ IoU thresholds \{0.5, 0.6, 0.7, 0.8\} to compute the AP, so as to prove that the STR module can produce more accurate localization. As shown in Table \ref{tab:aps}, the STR module produces consistently accurate detection results than the baseline method. The gap between the AP across all three subsets increases as the IoU threshold increases, which indicate that the STR module is important to produce more accurate detections. In addition, coupled with the STC module, the performance is further improved to $96.1\%$, $95.0\%$ and $90.1\%$ on the Easy, Medium and Hard subsets, respectively.

\vspace{-1.5mm}
\begin{table}[h]
\centering
\setlength{\tabcolsep}{3pt}
\caption{AP performance of the two-step regression applied to each pyramid level.}
\setlength{\tabcolsep}{5.2pt}
\begin{tabular}{c|c|cccccc}
\toprule[1.5pt]
STR & B & P2 & P3 & P4 & P5 & P6 & P7 \\
\hline
{\em Easy} & 95.1 & 94.8 & 94.3 & 94.8 & \bf 95.4 & \bf 95.7 & \bf 95.6 \\
{\em Medium} & 93.9 & 93.4 & 93.7 & 93.9 & \bf 94.2 & \bf 94.4 & \bf 94.6  \\
{\em Hard} & 88.0 & 87.5 & 87.7 & 87.0 & \bf 88.2 & \bf 88.2 & \bf 88.4 \\
\bottomrule[1.5pt]
\end{tabular}
\vspace{-5mm}
\label{tab:str_per_level}
\end{table}

\begin{table}[h]
\centering
\caption{AP at different IoU thresholds on the WIDER FACE Hard subset.}
\setlength{\tabcolsep}{10.5pt}
\begin{tabular}{c|ccccc}
\toprule[1.5pt]
IoU & 0.5 & 0.6 & 0.7 & 0.8 \\
\hline
{RetinaNet} & 88.1 & 76.4 & 57.8 & 28.5\\
{SRN (STR only)} & 88.8 & 83.4 & 66.5 & 38.2\\
\bottomrule[1.5pt]
\end{tabular}
\vspace{-0.5mm}
\label{tab:aps}
\end{table}

{\flushleft \textbf{Receptive Field Enhancement.} }
The RFE is used to diversify the receptive fields of detection layers in order to capture faces with extreme poses. Comparing the detection results between fourth and fifth columns in Table \ref{tab:ablation}, we notice that  RFE consistently improves the AP scores in different subsets, \ie, $0.3\%$, $0.3\%$, and $0.1\%$ APs on the Easy, Medium, and Hard categories. These improvements can be mainly attributed to the diverse receptive fields, which is useful to capture various pose faces for better detection accuracy.

\subsection{Evaluation on Benchmark}
{\flushleft \textbf{AFW Dataset.} }
It consists of $205$ images with $473$ labeled faces. The images in the dataset contains cluttered backgrounds with large variations in both face viewpoint and appearance. We compare SRN against seven state-of-the-art methods and three commercial face detectors (\ie, Face.com, Face++ and Picasa). As shown in Figure \ref{fig:AFW}, SRN outperforms these state-of-the-art methods with the top AP score ($99.87\%$).

{\flushleft \textbf{PASCAL Face Dataset.} }
It has $1,335$ labeled faces in $851$ images with large face appearance and pose variations. We present the precision-recall curves of the proposed SRN method and six state-of-the-art methods and three commercial face detectors (\ie, SkyBiometry, Face++ and Picasa) in Figure \ref{fig:PASCAL}. SRN achieves the state-of-the-art results by improving $4.99\%$ AP score compared to the second best method STN \cite{DBLP:conf/eccv/ChenHW016}.

{\flushleft \textbf{FDDB Dataset.} }
It contains $5,171$ faces annotated in $2,845$ images with a wide range of difficulties, such as occlusions, difficult poses, and low image resolutions. We evaluate the proposed SRN detector on the FDDB dataset and compare it with several state-of-the-art methods. As shown in Figure \ref{fig:FDDB}, our SRN sets a new state-of-the-art performance, \ie, $98.8\%$ true positive rate when the number of false positives is equal to $1000$. These results indicate that SRN is robust to varying scales, large appearance changes, heavy occlusions, and severe blur degradations that are prevalent in detecting face in unconstrained real-life scenarios.

{\flushleft \textbf{WIDER FACE Dataset.} }
We compare SRN with eighteen state-of-the-art face detection methods on both the validation and testing sets. To obtain the evaluation results on the testing set, we submit the detection results of SRN to the authors for evaluation. As shown in Figure \ref{fig:wider-face-ap}, we find that SRN performs favourably against the state-of-the-art based on the average precision (AP) across the three subsets, especially on the Hard subset which contains a large amount of small faces. Specifically, it produces the best AP scores in all subsets of both validation and testing sets, \ie, $96.4\%$ (Easy), $95.3\%$ (Medium) and $90.2\%$ (Hard) for validation set, and $95.9\%$ (Easy), $94.9\%$ (Medium) and $89.7\%$ (Hard) for testing set, surpassing all approaches, which demonstrates the superiority of the proposed detector.

\section{Conclusion}
In this paper, we have presented SRN, a novel single shot face detector, which consists of two key modules, \ie, the STC and the STR. The STC uses the first-step classifier to filter out most simple negative anchors from low level detection layers to reduce the search space for the second-step classifier, so as to reduce false positives. And the STR applies the first-step regressor to coarsely adjust the locations and sizes of anchors from high level detection layers to provide better initialization for the second-step regressor, in order to improve the location accuracy of bounding boxes. Moreover, the RFE is introduced to provide diverse receptive fields to better capture faces in some extreme poses. Extensive experiments on the AFW, PASCAL face, FDDB and WIDER FACE datasets demonstrate that SRN achieves the state-of-the-art detection performance.

\clearpage
\small
\bibliographystyle{aaai}
\bibliography{reference}

\begin{thebibliography}{}

\bibitem[\protect\citeauthoryear{Bai \bgroup et al\mbox.\egroup
  }{2018}]{bai2018finding}
Bai, Y.; Zhang, Y.; Ding, M.; and Ghanem, B.
\newblock 2018.
\newblock Finding tiny faces in the wild with generative adversarial network.
\newblock In {\em CVPR}.

\bibitem[\protect\citeauthoryear{Brubaker \bgroup et al\mbox.\egroup
  }{2008}]{DBLP:journals/ijcv/BrubakerWSMR08}
Brubaker, S.~C.; Wu, J.; Sun, J.; Mullin, M.~D.; and Rehg, J.~M.
\newblock 2008.
\newblock On the design of cascades of boosted ensembles for face detection.
\newblock {\em IJCV}.

\bibitem[\protect\citeauthoryear{Cai and
  Vasconcelos}{2018}]{DBLP:journals/corr/abs-1712-00726}
Cai, Z., and Vasconcelos, N.
\newblock 2018.
\newblock Cascade {R-CNN:} delving into high quality object detection.
\newblock In {\em CVPR}.

\bibitem[\protect\citeauthoryear{Chen \bgroup et al\mbox.\egroup
  }{2016}]{DBLP:conf/eccv/ChenHW016}
Chen, D.; Hua, G.; Wen, F.; and Sun, J.
\newblock 2016.
\newblock Supervised transformer network for efficient face detection.
\newblock In {\em ECCV}.

\bibitem[\protect\citeauthoryear{Gidaris and
  Komodakis}{2015}]{DBLP:conf/iccv/GidarisK15}
Gidaris, S., and Komodakis, N.
\newblock 2015.
\newblock Object detection via a multi-region and semantic segmentation-aware
  {CNN} model.
\newblock In {\em ICCV}.

\bibitem[\protect\citeauthoryear{Girshick}{2015}]{DBLP:conf/iccv/Girshick15}
Girshick, R.~B.
\newblock 2015.
\newblock Fast {R-CNN}.
\newblock In {\em ICCV}.

\bibitem[\protect\citeauthoryear{Hao \bgroup et al\mbox.\egroup
  }{2017}]{hao2017scale}
Hao, Z.; Liu, Y.; Qin, H.; Yan, J.; Li, X.; and Hu, X.
\newblock 2017.
\newblock Scale-aware face detection.
\newblock In {\em CVPR}.

\bibitem[\protect\citeauthoryear{He \bgroup et al\mbox.\egroup
  }{2016}]{DBLP:conf/cvpr/HeZRS16}
He, K.; Zhang, X.; Ren, S.; and Sun, J.
\newblock 2016.
\newblock Deep residual learning for image recognition.
\newblock In {\em CVPR}.

\bibitem[\protect\citeauthoryear{Howard}{2013}]{DBLP:journals/corr/Howard13}
Howard, A.~G.
\newblock 2013.
\newblock Some improvements on deep convolutional neural network based image
  classification.
\newblock {\em CoRR}.

\bibitem[\protect\citeauthoryear{Hu and Ramanan}{2017}]{DBLP:conf/cvpr/HuR17}
Hu, P., and Ramanan, D.
\newblock 2017.
\newblock Finding tiny faces.
\newblock In {\em CVPR}.

\bibitem[\protect\citeauthoryear{Jain and Learned-Miller}{2010}]{fddbTech}
Jain, V., and Learned-Miller, E.
\newblock 2010.
\newblock Fddb: A benchmark for face detection in unconstrained settings.
\newblock Technical report, University of Massachusetts, Amherst.

\bibitem[\protect\citeauthoryear{Li \bgroup et al\mbox.\egroup
  }{2015}]{DBLP:conf/cvpr/LiLSBH15}
Li, H.; Lin, Z.; Shen, X.; Brandt, J.; and Hua, G.
\newblock 2015.
\newblock A convolutional neural network cascade for face detection.
\newblock In {\em CVPR}.

\bibitem[\protect\citeauthoryear{Liao, Jain, and
  Li}{2016}]{DBLP:journals/pami/LiaoJL16}
Liao, S.; Jain, A.~K.; and Li, S.~Z.
\newblock 2016.
\newblock A fast and accurate unconstrained face detector.
\newblock {\em TPAMI}.

\bibitem[\protect\citeauthoryear{Lin \bgroup et al\mbox.\egroup
  }{2014}]{DBLP:conf/eccv/LinMBHPRDZ14}
Lin, T.; Maire, M.; Belongie, S.~J.; Hays, J.; Perona, P.; Ramanan, D.;
  Doll{\'{a}}r, P.; and Zitnick, C.~L.
\newblock 2014.
\newblock Microsoft {COCO:} common objects in context.
\newblock In {\em ECCV}.

\bibitem[\protect\citeauthoryear{Lin \bgroup et al\mbox.\egroup
  }{2017a}]{DBLP:conf/cvpr/LinDGHHB17}
Lin, T.; Doll{\'{a}}r, P.; Girshick, R.~B.; He, K.; Hariharan, B.; and
  Belongie, S.~J.
\newblock 2017a.
\newblock Feature pyramid networks for object detection.
\newblock In {\em CVPR}.

\bibitem[\protect\citeauthoryear{Lin \bgroup et al\mbox.\egroup
  }{2017b}]{DBLP:conf/iccv/LinPRK17}
Lin, T.; Goyal, P.; Girshick, R.~B.; He, K.; and Doll{\'{a}}r, P.
\newblock 2017b.
\newblock Focal loss for dense object detection.
\newblock In {\em ICCV}.

\bibitem[\protect\citeauthoryear{Liu \bgroup et al\mbox.\egroup
  }{2016}]{DBLP:conf/eccv/LiuAESRFB16}
Liu, W.; Anguelov, D.; Erhan, D.; Szegedy, C.; Reed, S.~E.; Fu, C.; and Berg,
  A.~C.
\newblock 2016.
\newblock {SSD:} single shot multibox detector.
\newblock In {\em ECCV}.

\bibitem[\protect\citeauthoryear{Mathias \bgroup et al\mbox.\egroup
  }{2014}]{DBLP:conf/eccv/MathiasBPG14}
Mathias, M.; Benenson, R.; Pedersoli, M.; and Gool, L. J.~V.
\newblock 2014.
\newblock Face detection without bells and whistles.
\newblock In {\em ECCV}.

\bibitem[\protect\citeauthoryear{Najibi \bgroup et al\mbox.\egroup
  }{2017}]{DBLP:conf/iccv/NajibiSCD17}
Najibi, M.; Samangouei, P.; Chellappa, R.; and Davis, L.~S.
\newblock 2017.
\newblock {SSH:} single stage headless face detector.
\newblock In {\em ICCV}.

\bibitem[\protect\citeauthoryear{Paszke \bgroup et al\mbox.\egroup
  }{2017}]{paszke2017pytorch}
Paszke, A.; Gross, S.; Chintala, S.; and Chanan, G.
\newblock 2017.
\newblock Pytorch.

\bibitem[\protect\citeauthoryear{Pham and Cham}{2007}]{DBLP:conf/iccv/PhamC07}
Pham, M., and Cham, T.
\newblock 2007.
\newblock Fast training and selection of haar features using statistics in
  boosting-based face detection.
\newblock In {\em ICCV}.

\bibitem[\protect\citeauthoryear{Qin \bgroup et al\mbox.\egroup
  }{2016}]{DBLP:conf/cvpr/QinYLH16}
Qin, H.; Yan, J.; Li, X.; and Hu, X.
\newblock 2016.
\newblock Joint training of cascaded {CNN} for face detection.
\newblock In {\em CVPR}.

\bibitem[\protect\citeauthoryear{Ren \bgroup et al\mbox.\egroup
  }{2017}]{DBLP:journals/pami/RenHG017}
Ren, S.; He, K.; Girshick, R.~B.; and Sun, J.
\newblock 2017.
\newblock Faster {R-CNN:} towards real-time object detection with region
  proposal networks.
\newblock {\em TPAMI}.

\bibitem[\protect\citeauthoryear{Russakovsky \bgroup et al\mbox.\egroup
  }{2015}]{DBLP:journals/ijcv/RussakovskyDSKS15}
Russakovsky, O.; Deng, J.; Su, H.; Krause, J.; Satheesh, S.; Ma, S.; Huang, Z.;
  Karpathy, A.; Khosla, A.; Bernstein, M.~S.; Berg, A.~C.; and Li, F.
\newblock 2015.
\newblock Imagenet large scale visual recognition challenge.
\newblock {\em IJCV}.

\bibitem[\protect\citeauthoryear{Shi \bgroup et al\mbox.\egroup
  }{2018}]{shi2018real}
Shi, X.; Shan, S.; Kan, M.; Wu, S.; and Chen, X.
\newblock 2018.
\newblock Real-time rotation-invariant face detection with progressive
  calibration networks.
\newblock In {\em CVPR}.

\bibitem[\protect\citeauthoryear{Song \bgroup et al\mbox.\egroup
  }{2018}]{song2018beyond}
Song, G.; Liu, Y.; Jiang, M.; Wang, Y.; Yan, J.; and Leng, B.
\newblock 2018.
\newblock Beyond trade-off: Accelerate fcn-based face detector with higher
  accuracy.
\newblock In {\em CVPR}.

\bibitem[\protect\citeauthoryear{Szegedy \bgroup et al\mbox.\egroup
  }{2015}]{DBLP:conf/cvpr/SzegedyLJSRAEVR15}
Szegedy, C.; Liu, W.; Jia, Y.; Sermanet, P.; Reed, S.~E.; Anguelov, D.; Erhan,
  D.; Vanhoucke, V.; and Rabinovich, A.
\newblock 2015.
\newblock Going deeper with convolutions.
\newblock In {\em CVPR}.

\bibitem[\protect\citeauthoryear{Tang \bgroup et al\mbox.\egroup
  }{2018}]{tang2018pyramidbox}
Tang, X.; Du, D.~K.; He, Z.; and Liu, J.
\newblock 2018.
\newblock Pyramidbox: A context-assisted single shot face detector.
\newblock In {\em ECCV}.

\bibitem[\protect\citeauthoryear{Viola and
  Jones}{2004}]{DBLP:journals/ijcv/ViolaJ04}
Viola, P.~A., and Jones, M.~J.
\newblock 2004.
\newblock Robust real-time face detection.
\newblock {\em IJCV}.

\bibitem[\protect\citeauthoryear{Wang \bgroup et al\mbox.\egroup
  }{2017a}]{wang2017face}
Wang, H.; Li, Z.; Ji, X.; and Wang, Y.
\newblock 2017a.
\newblock Face r-cnn.
\newblock {\em CoRR}.

\bibitem[\protect\citeauthoryear{Wang \bgroup et al\mbox.\egroup
  }{2017b}]{wang2017detecting}
Wang, Y.; Ji, X.; Zhou, Z.; Wang, H.; and Li, Z.
\newblock 2017b.
\newblock Detecting faces using region-based fully convolutional networks.
\newblock {\em CoRR}.

\bibitem[\protect\citeauthoryear{Wang, Yuan, and Yu}{2017}]{wang2017fan}
Wang, J.; Yuan, Y.; and Yu, G.
\newblock 2017.
\newblock Face attention network: An effective face detector for the occluded
  faces.
\newblock {\em CoRR}.

\bibitem[\protect\citeauthoryear{Yan \bgroup et al\mbox.\egroup
  }{2014a}]{DBLP:conf/cvpr/YanLWL14}
Yan, J.; Lei, Z.; Wen, L.; and Li, S.~Z.
\newblock 2014a.
\newblock The fastest deformable part model for object detection.
\newblock In {\em CVPR}.

\bibitem[\protect\citeauthoryear{Yan \bgroup et al\mbox.\egroup
  }{2014b}]{DBLP:journals/ivc/YanZLL14}
Yan, J.; Zhang, X.; Lei, Z.; and Li, S.~Z.
\newblock 2014b.
\newblock Face detection by structural models.
\newblock {\em IVC}.

\bibitem[\protect\citeauthoryear{Yang \bgroup et al\mbox.\egroup
  }{2015}]{DBLP:conf/iccv/YangLLT15}
Yang, S.; Luo, P.; Loy, C.~C.; and Tang, X.
\newblock 2015.
\newblock From facial parts responses to face detection: {A} deep learning
  approach.
\newblock In {\em ICCV}.

\bibitem[\protect\citeauthoryear{Yang \bgroup et al\mbox.\egroup
  }{2016}]{DBLP:conf/cvpr/YangLLT16}
Yang, S.; Luo, P.; Loy, C.~C.; and Tang, X.
\newblock 2016.
\newblock {WIDER} {FACE:} {A} face detection benchmark.
\newblock In {\em CVPR}.

\bibitem[\protect\citeauthoryear{Yu \bgroup et al\mbox.\egroup
  }{2016}]{DBLP:conf/mm/YuJWCH16}
Yu, J.; Jiang, Y.; Wang, Z.; Cao, Z.; and Huang, T.~S.
\newblock 2016.
\newblock Unitbox: An advanced object detection network.
\newblock In {\em ACMMM}.

\bibitem[\protect\citeauthoryear{Zhang \bgroup et al\mbox.\egroup
  }{2016}]{DBLP:journals/spl/ZhangZLQ16}
Zhang, K.; Zhang, Z.; Li, Z.; and Qiao, Y.
\newblock 2016.
\newblock Joint face detection and alignment using multitask cascaded
  convolutional networks.
\newblock {\em SPL}.

\bibitem[\protect\citeauthoryear{Zhang \bgroup et al\mbox.\egroup
  }{2017}]{DBLP:conf/iccv/abs-1708-05237}
Zhang, S.; Zhu, X.; Lei, Z.; Shi, H.; Wang, X.; and Li, S.~Z.
\newblock 2017.
\newblock S\({}^{\mbox{3}}\){FD}: Single shot scale-invariant face detector.
\newblock In {\em ICCV}.

\bibitem[\protect\citeauthoryear{Zhang \bgroup et al\mbox.\egroup
  }{2018}]{DBLP:journals/corr/abs-1711-06897}
Zhang, S.; Wen, L.; Bian, X.; Lei, Z.; and Li, S.~Z.
\newblock 2018.
\newblock Single-shot refinement neural network for object detection.
\newblock In {\em CVPR}.

\bibitem[\protect\citeauthoryear{Zhu and Ramanan}{2012}]{DBLP:conf/cvpr/ZhuR12}
Zhu, X., and Ramanan, D.
\newblock 2012.
\newblock Face detection, pose estimation, and landmark localization in the
  wild.
\newblock In {\em CVPR}.

\bibitem[\protect\citeauthoryear{Zhu \bgroup et al\mbox.\egroup
  }{2018}]{zhu2018seeing}
Zhu, C.; Tao, R.; Luu, K.; and Savvides, M.
\newblock 2018.
\newblock Seeing small faces from robust anchor’s perspective.
\newblock In {\em CVPR}.

\bibitem[\protect\citeauthoryear{Zitnick and
  Doll{\'{a}}r}{2014}]{DBLP:conf/eccv/ZitnickD14}
Zitnick, C.~L., and Doll{\'{a}}r, P.
\newblock 2014.
\newblock Edge boxes: Locating object proposals from edges.
\newblock In {\em ECCV}.

\end{thebibliography}
\end{document}